\documentclass[nohyperref]{article}

\usepackage{microtype}
\usepackage{graphicx}
\usepackage{subfigure}
\usepackage{booktabs}

\usepackage{siunitx}  

\usepackage[accepted]{icml2022}      

\usepackage{amsmath}
\usepackage{amssymb}
\usepackage{mathtools}
\usepackage{amsthm}
\usepackage{algorithm}
\usepackage{algorithmic}
\usepackage{graphicx}
\usepackage{color}

\floatname{algorithm}{Algorithm}
\renewcommand{\algorithmicrequire}{\textbf{Input:}}
\renewcommand{\algorithmicensure}{\textbf{Output:}}

\usepackage[pagebackref,breaklinks,colorlinks]{hyperref}

\usepackage[capitalize]{cleveref}
\crefname{section}{Sec.}{Secs.}
\Crefname{section}{Section}{Sections}
\Crefname{table}{Table}{Tables}
\crefname{table}{Tab.}{Tabs.}

\theoremstyle{plain}

\theoremstyle{definition}

\theoremstyle{remark}

\usepackage[textsize=tiny]{todonotes}

\icmltitlerunning{GradTail: Learning Long-Tailed Data Using Gradient-based Sample Weighting}

\begin{document}

\twocolumn[
\icmltitle{GradTail: Learning Long-Tailed Data Using Gradient-based Sample Weighting}

\icmlsetsymbol{equal}{*}

\begin{icmlauthorlist}
\icmlauthor{Zhao Chen}{waymo}
\icmlauthor{Vincent Casser}{waymo}
\icmlauthor{Henrik Kretzschmar}{waymo}
\icmlauthor{Dragomir Anguelov}{waymo}
\end{icmlauthorlist}

\icmlaffiliation{waymo}{Waymo LLC, Mountain View, California, USA}

\icmlcorrespondingauthor{Zhao Chen}{zhaoch@waymo.com}

\icmlkeywords{Machine Learning, ICML}

\vskip 0.3in
]
  
\printAffiliationsAndNotice{}

\begin{abstract}
   We propose GradTail, an algorithm that uses gradients to improve model performance on the fly in the face of long-tailed training data distributions. Unlike conventional long-tail classifiers which operate on converged - and possibly overfit - models, we demonstrate that an approach based on gradient dot product agreement can isolate long-tailed data early on during model training and improve performance by dynamically picking higher sample weights for that data. We show that such upweighting leads to model improvements for both classification and regression models, the latter of which are relatively unexplored in the long-tail literature, and that the long-tail examples found by gradient alignment are consistent with our semantic expectations. 
\end{abstract}

\section{Introduction}
\label{sec:intro}

Although modern deep learning and machine learning techniques have achieved impressive performance across a diverse set of tasks, most models still struggle when faced with uneven data distributions containing very rare or long-tail examples. Such rare examples are frequently present in real world data \cite{bengio2015battle}, and handling them well is crucial in safety-critical applications like autonomous driving (e.g. \cite{philion2019fastdraw}). 

Conventional methods to mitigate the effects of long-tailed training distributions often require identifying long-tailed data and then labeling additional data of the same type (i.e. active learning), like in \cite{activelearning1}, which in effect increases the probability density of such data and moves it out of the long-tail. However, such data-driven approaches are expensive and may not ever definitively solve the problem, as squashing one set of long-tail examples can often lead to the formation of others.

We focus within this work not on collecting more data, but rather dynamically upweighting long-tailed examples during training. At the core of this line of research is the assumption that the long-tailedness of a particular example is not only a function of the data distribution, but also the state of the model itself. Examples that are at one point long-tailed may be learned properly through dynamic upweighting and become more in-distribution later in training. 

Such a setting is relatively unexplored within the long-tail context; to wit, the majority of long-tail classification techniques, such as entropy \cite{entropy1} or ensembling \cite{ensemble1}, prove unsuitable for dynamic upweighting as they rely on model convergence to derive meaningful uncertainty signals. In contrast, our work rests on the claim that there is already rich information available well before a model converges, and that tapping into that information allows us to mitigate long-tail effects on the fly.

Using model dynamics as our long-tail probe allows us to tackle another fundamental issue in long-tail learning: how do we differentiate examples that are properly in the long-tail versus examples that are purely difficult? More formally, long-tail examples have high \textit{reducible, epistemic uncertainty}, as a model can in principle learn them but struggles to (in this case, because they are rare). In contrast, those examples that exhibit high \textit{irreducible, aleatoric uncertainty} are hard but \textit{not} long-tail, as their difficulty derives from more fundamental sources of noise within the data rather than pure rarity \cite{kendall2017uncertainties}. Visual occlusions are potentially good examples of the latter category, as fully or mostly occluded objects cannot be detected by many vision systems regardless of how many examples of them exist in the training dataset. For the rest of this work, we will refer to examples of high aleatoric uncertainty as ``hard," and examples of high epistemic uncertainty as ``rare." We use quote-marks here to emphasize that although these labels are theoretically grounded, they deviate somewhat from what ``rare" generally means in the literature. A discussion of this discrepancy will be given in Section \ref{sec:toy-example}, and more discussion is added in Appendix \ref{sec:uncertainty-discussion}.

\begin{figure*}[ht]
\centering
\includegraphics[width=0.85\linewidth]{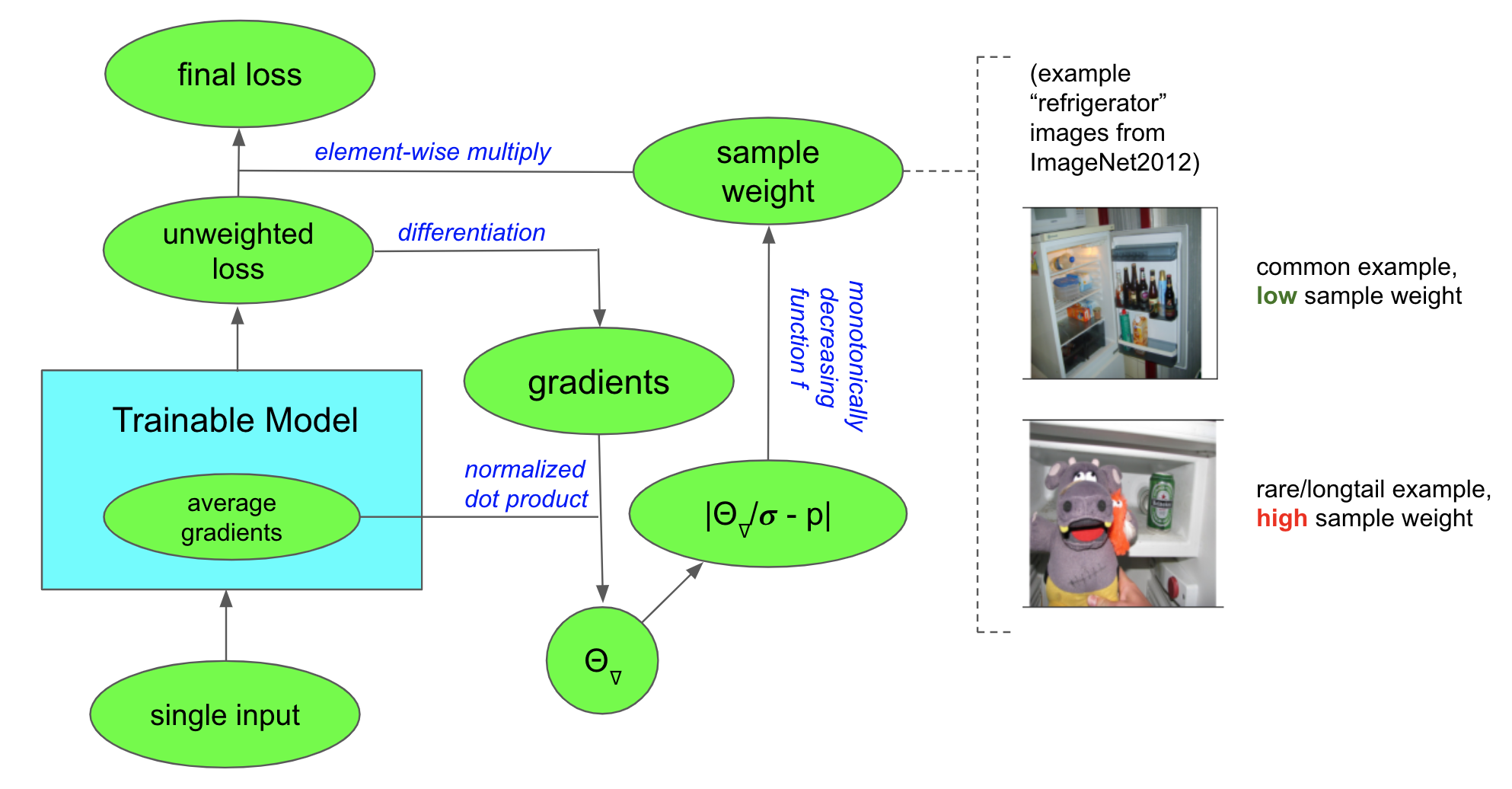}
\caption{Schematic description of the proposed GradTail technique. Gradients (with respect to some trainable weights) are taken from an unweighted loss function and compared to an average gradient vector. The resultant metric is normalized and a distance to hyperparameter pivot $p$ is computed. This distance is then converted to a sample weight via a monotonic function which produces the final weighted loss. A choice of $p$ near zero ensures that examples upweighted by GradTail are long-tail (i.e. contain high epistemic uncertainty).}
\label{fig:gradtail_schematic}
\end{figure*}

As we are focusing on system dynamics to mitigate long-tailed data, it is natural for us to hone in on \textit{gradients} as the core entities with which to perform our calculations. Given two loss gradients $\nabla_wL(x_0; w), \nabla_wL(x_1; w)$ for loss $L$, trainable examples $x_0, x_1$ and trainable weight $w$, the simple dot product $\lambda\nabla_wL(x_0) \cdot \nabla_wL(x_1)$ tells us the change in $L(x_0)$ should we follow the gradient $\nabla_wL(x_1)$ for a step size of $\lambda$ (or, symmetrically, the change in $L(x_1)$ upon an update in direction $\lambda\nabla_wL(x_0)$. Crucially, we can calculate the \textit{average gradient vector}, $E_x[\nabla_wL(x)]$ and the dot product $\lambda\nabla_wL(x_0) \cdot E_x[\nabla_wL(x)]$ tells us how any particular example is affected by an update for the mean example within a distribution. We note that this gradient quantity has a number of desirable properties that make it especially suited for our purposes:

\begin{enumerate}
    \item \textbf{Dynamic}: Gradients can be calculated on the fly, and in fact are already calculated as part of the standard backpropagation loop.
    \item \textbf{Efficient}: Because gradients are already calculated during normal training, our method has low compute overhead.
    \item \textbf{Explicit comparison}: The gradient dot product is an \textit{explicit} comparison between each example and the mean of a distribution, as opposed to the implicit comparison that many uncertainty methods use.
    \item \textbf{Separability of uncertainty}: Gradients can separate whether a problematic example is truly rare (and thus still learnable) or just hard. Examples that backpropagate nearly orthogonal gradients, for instance, are in principle learnable by the model because application of that gradient will not obstruct learning of the average example. Examples that backpropagate opposing gradients may indicate overly noisy data or incompatibility with the model, and thus may just be hard.
    \item \textbf{Nondependence on convergence}: Gradients are discriminative early on in the training process, and do not require a converged model to be taken as a useful signal. 
    \item \textbf{Dependence on labels}: Gradients are explicitly dependent on labels. Thus, not only are they sensitive to out-of-distribution inputs, they are also sensitive to corrupted labels and other unwanted label behaviors. In contrast, many standard methods for uncertainty estimation are only explicitly dependent on the input. 
\end{enumerate}

In general, the separability property posited in (4) allows us to isolate examples that are difficult but still learnable by the model. We hypothesize that by emphasizing these examples, we allow the model to explore parts of parameter space that are benign but which would be left untouched otherwise by the model.

Our main contributions are as follows: 

\begin{itemize}
    \item We introduce GradTail, an algorithm that can be implemented on the fly during model training to improve overall performance with emphasis on data in the long-tail.
    \item We show through experiments with controlled synthetic data how gradient dot products produce a natural continuous scale to separate difficult (aleatoric) data from rare (epistemic) data.
    \item We demonstrate that GradTail works well in traditional sample weighting settings such as classification as well as in dense regression settings, which are often overlooked by the sample weighting literature.
\end{itemize}

\section{Related Work}

\textbf{Uncertainty estimation} is of keen interest to deep learning practitioners, as it provides useful information about failure modes and potential pathways towards improvement for a given model. Much of the work within uncertainty estimation is related to work on Bayesian deep learning \cite{bayes1, bayes3, bayes2}, in which distributions placed over network weights or inputs provide explicit measures of predictive uncertainty. Extending model training to explicitly model uncertainties can also lead to more robust training outputs \cite{kendall2017uncertainties}. Modeling uncertainties is closely related to active learning \cite{activelearning1, activelearning2}, where uncertainty measurements can be used to collect new labeled data that will maximally benefit the model.

\textbf{Example-level weighting} methods have been well-studied \cite{imbalancedata}, and can even be done on the fly \cite{lin2017focal}. Reweighting has also become popular in multitask learning \cite{mtlweight1, mtlweight2}, where different tasks must be balanced with each other for optimal training. Multitask learning also has popularized gradient comparison techniques \cite{mtlgrad1, mtlgrad2}, which we leverage heavily within this current work.

\textbf{Out-of-distribution and long-tail detection} provide important tools to classify and mitigate the effect of data that lie far from the main data manifold. Entropy \cite{entropy1, entropy2} and ensemble \cite{ensemble1, ensemble2} methods can select for the tails of our data distribution, and various learning methods exist to mitigate their effect on model performance \cite{longtailmethod1, longtailmethod2}. However, these methods usually focus on classification and are explicitly coupled with semantically defined long-tails within curated datasets (e.g. \cite{dataset1}). We instead seek a method that can operate without any long-tail information provided in the training data. 
\section{Methodology}

We present the main algorithm loop for GradTail dynamic upweighting in Algorithm \ref{alg:gradtail}. We note that the actual algorithm is very simple and only involves standard dot product and arithmetic operations. A schematic of the GradTail method is shown in Figure \ref{fig:gradtail_schematic}.

\begin{algorithm}
\caption{Gradient Long-Tail Dynamic Upweighting (GradTail)}\label{alg:gradtail}
\begin{algorithmic}
\vspace{0.07in}
\STATE \textbf{note} that all dot products denoted by $\cdot$ are normalized and therefore lie in the range $[-1, 1]$.
\STATE \textbf{note} that all operations keep the batch dimension intact unless explicitly noted.\\ \vspace{-0.03in}
\hrulefill
\vspace{0.01in}
\STATE \textbf{choose} monotonically decreasing and positive activation function $f:[0,\infty]\mapsto [1,\infty]$
\STATE \textbf{choose} subset of trainable weights $\mathbf{w} \underline{\subset} W$ from model $\Phi$.
\STATE \textbf{choose} dataset $(\mathbf{X}, \mathbf{Y})$ with minibatches $\mathbf{x}\underline{\subset}\mathbf{X}$, $\mathbf{y}\underline{\subset}\mathbf{Y}$.
\STATE \textbf{choose} loss function $L$ to calculate gradients.
\STATE \textbf{choose} pivot $p$ and decay $\lambda$.\\
\STATE \textbf{initialize} to zeroes $\tilde{\mathbf{w}}$, a tensor the same shape as $\mathbf{w}$ that will hold the average gradient.
\STATE \textbf{initialize} to zero average variance variable $\sigma$.

\item[]
\FUNCTION{GetLossForMinibatch($\mathbf{x}$, $\mathbf{y}$)}
\STATE \textbf{calculate} $\nabla(\mathbf{x})$ := $\nabla_{\mathbf{w}} L(\Phi(\mathbf{x}), \mathbf{y})$
\STATE \textbf{calculate} $\theta(\mathbf{x}) = \nabla(\mathbf{x})\cdot \tilde{\mathbf{w}}$
\STATE \textbf{calculate} $\sigma_{\mathbf{x}} = E_{\mathbf{x}}[|\theta(\mathbf{x})|]$
\STATE \textbf{set} $\sigma = \lambda\sigma + (1-\lambda) \sigma_{\mathbf{x}}$
\STATE \textbf{set} $\tilde{\mathbf{w}} = \lambda \tilde{\mathbf{w}} + (1-\lambda)E_{\mathbf{x}}[\nabla(\mathbf{x})]$
\STATE \textbf{calculate} loss weights $\mathbf{q} = f(|\frac{\theta(\mathbf{x})}{\sigma}-p|)$\\
\STATE \textbf{return} {$\sum \mathbf{q}L(\Phi(\mathbf{x}), \mathbf{y})$}
\ENDFUNCTION
\end{algorithmic}
\end{algorithm}

The main hyperparameter is the pivot $p$, which presents us a lever to tune the tradeoff between exploration and sensitivity to aleatoric uncertainty. A more negative pivot means that we upweight examples that are less in agreement with the mean sample gradient, but for milder negative values can allow the model's trainable parameter space to explore in relatively benign directions. Being able to control this tradeoff is a key feature of GradTail, as it allows us to filter out examples that are outliers and incompatible with the bulk of the data distribution. In general, selecting a pivot of 0 provides a good baseline. 

Another important hyperparameter is the activation function $f$. $f$ is a monotonically decreasing function, which produces positive outputs that are $\geq 1$. $f$ is monotonically decreasing because it takes as input the absolute distance of the gradient dot product from some pivot point. The further the absolute distance from this pivot point, the less we want to upweight this particular sample. 

In our experiments within this work, we pick a sigmoid activation function $f(x) = 1.0 + \frac{A}{1-e^{-Bx}}$, where $A,B$ are additional hyperparameters. We pick this function as it presents relatively mild slopes but also has highest variance near $x=0$, which allows it to be especially peaked around gradient dot products near the pivot point. 
\section{Experiments}
We present results within this section on a number of experimental settings. We begin with in-depth discussion on how gradients are discriminative towards the long-tails of the distribution on a simple 2-dimensional synthetic example. We then move to real datasets and show strong results on both ImageNet2012 \cite{imagenet} and the monocular depth prediction task on the Waymo Open Dataset \cite{wod}. Exact details for models used within these experiments are provided within the Appendix. 

\subsection{A Simple Toy Example}
\label{sec:toy-example}
Before we test our methodology on large-scale settings, in which the long-tail is extremely high-dimensional and often semantically ambiguous, we find it instructive to see how gradients can help identify the long-tail of a data distribution within a well-understood low dimensional setting. 

\begin{figure}[ht]
\centering
    \subfigure[Ground Truth]{
      \centering
      \includegraphics[width=0.5\linewidth]{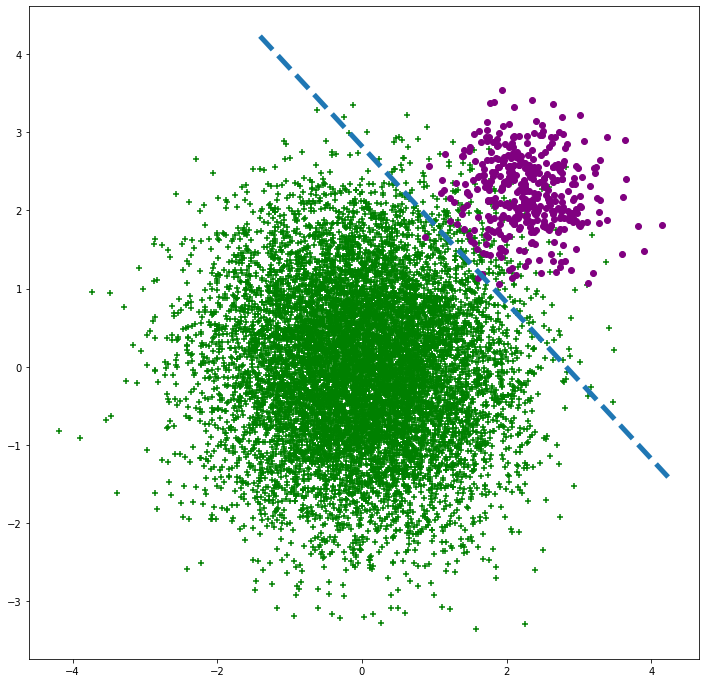}
    }
    \linebreak
    \subfigure[Base Model]{
      \centering
      \includegraphics[width=0.33\linewidth]{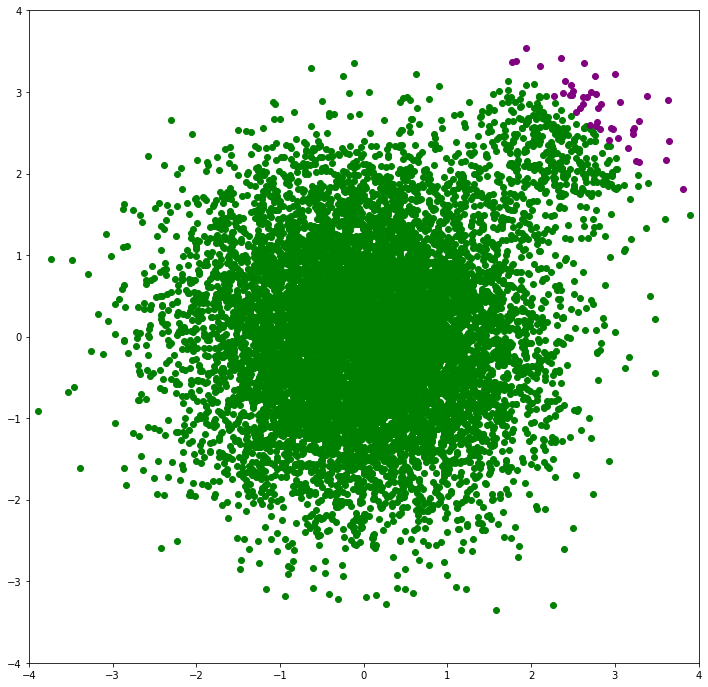}
    }
    \subfigure[GradTail]{
      \centering
      \includegraphics[width=0.33\linewidth]{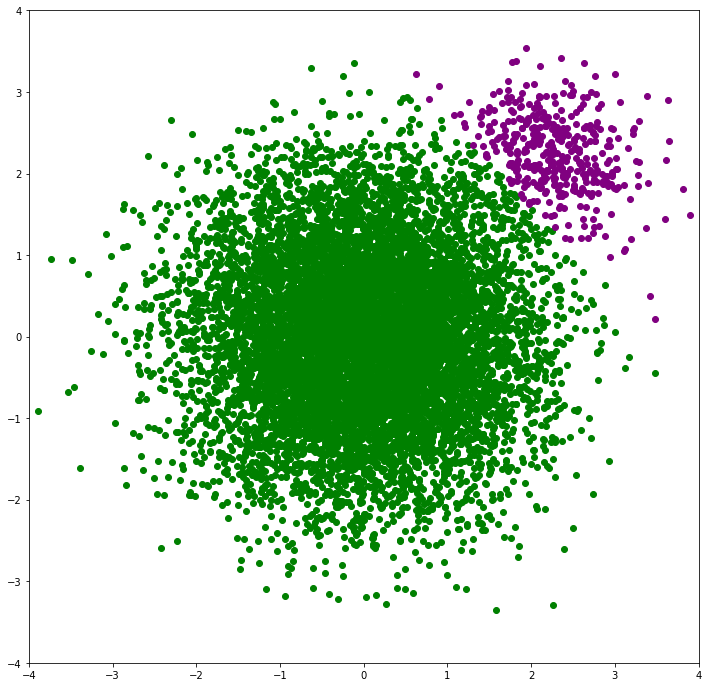}
    }
\caption{Toy example results for GradTail dynamic weighting. (a) Distribution of data within a 2-class toy example classifier. The common (green cross) class consists of 10,000 data points and follows a $\mathcal{N}(\mathbf{0}, \mathbf{I})$ distribution. The rare (purple circle) class consists of 400 data points and follows a $\mathcal{N}([2.2, 2.2], 0.5\mathbf{I})$ distribution. The dotted line shows the analytically correct decision boundary, calculated for when the two data generating distributions have equal probability. (b) Result when a baseline MLP classifier is trained on the data. The result shown here represents a better-than-median outcome, as most baseline models converge to the majority classifier. (c) Result of an MLP classifier when GradTail is applied during training.}
\label{fig:toy-example}
\end{figure}

For our toy example, we use a 2-class classifier for data in 2 dimensions. The data we use for much of our toy example analysis is displayed in Figure \ref{fig:toy-example}. As seen, the data consists of 10,000 common-class points (the green cross data) and 400 uncommon-class points (the purple circle data). The common class is distributed I.I.D. as the standard normal, while the uncommon class is right-upwards shifted to $\mathcal{N}([2.2,2.2], 0.5\mathbf{I})$ so that it is well-separated enough for the classification problem to be well-defined, while still close enough to the origin to create a difficult decision boundary. Our model is a simple 2-layer MLP with one hidden layer of five neurons. 

As expected, such a simple model often converges to the majority classifier of the common class. Even in the best case scenario the result of a naive MLP classifier looks something like the predictions in Figure \ref{fig:toy-example}(b), with the decision boundary pushed deep into the purple uncommon class's distribution density. Such a result is expected due to the lopsidedness of the class frequencies. However, as shown in Figure \ref{fig:toy-example}(c), training with GradTail results in a much more accurate decision boundary, which comes quite close to the ground truth decision boundary shown by the dotted line in Figure \ref{fig:toy-example}(a).

\begin{figure}[ht]
\centering
    \subfigure[GradTail]{
      \centering
      \includegraphics[width=0.3\linewidth]{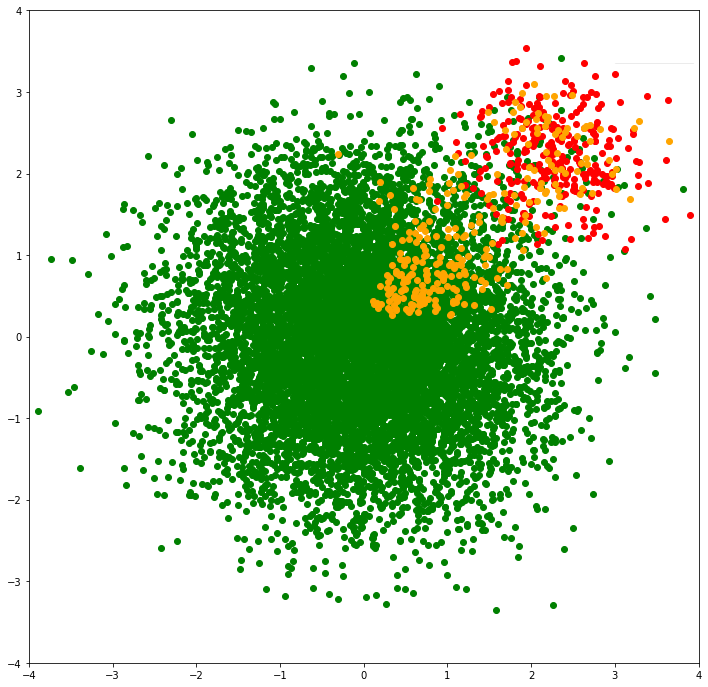}
    }
    \subfigure[GradTail (rare)]{
      \centering
      \includegraphics[width=0.3\linewidth]{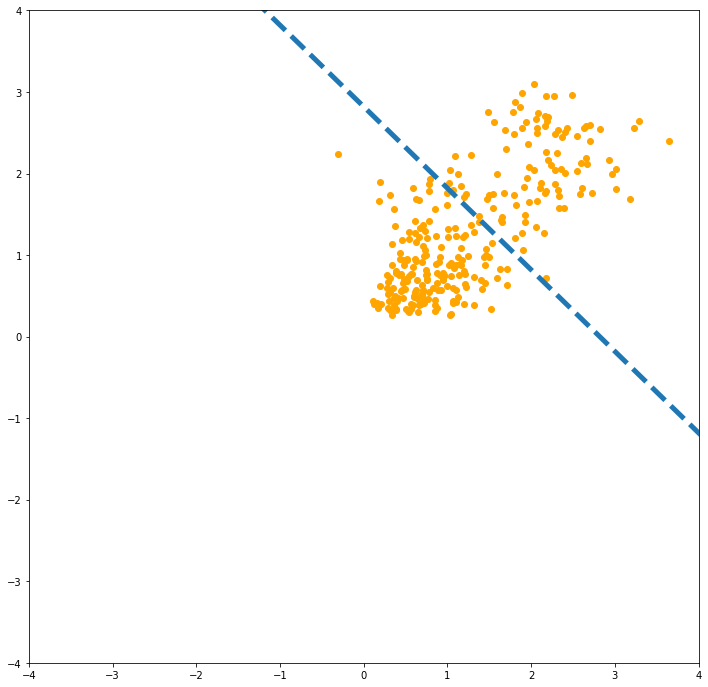}
    }
    \subfigure[Entropy]{
      \centering
      \includegraphics[width=0.3\linewidth]{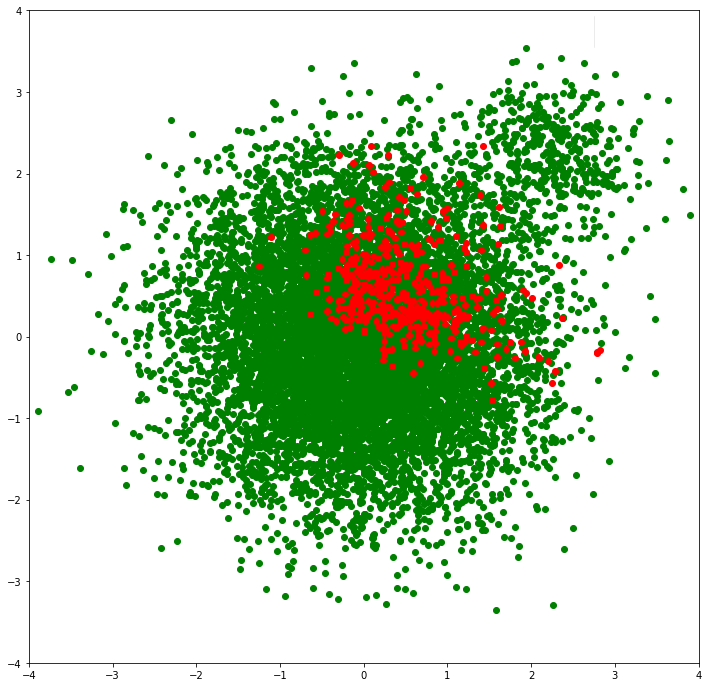}
    }
\caption{Toy dataset splits based on GradTail and other methods. We show how the toy dataset shown in Figure \ref{fig:toy-example} can be discretely clustered based on various uncertainty metrics. These classifications are based on aggregated statistics for each data point across the entire training run. (a) For GradTail, because our pivot value is around 0, we use a small closed-interval range around zero of $[-0.07, 0.07]$ as the range that we associate with rare examples that we aim to upweight. Examples with dot products under that range are labeled as ``hard" (red) examples, while examples with dot products above that range are labeled as ``common" (green). Many examples in the low-frequency class are observed to be ``hard," while both classes have examples that are labeled as ``rare" (yellow). (b) Only the ``rare" examples found by GradTail along with the true decision boundary as reference. The ``rare" examples are distributed evenly around the decision boundary, and thus upweighting these examples will lead to better delineation of the decision boundary. (c) High entropy and low entropy points are labeled in red and green, respectively. Because of the lopsided nature of the dataset, entropy proves to be a poor, incoherent predictor of data uncertainty, and the results look random.}
\label{fig:toy_example_class}
\end{figure}

To dive deeper into why GradTail performs better in this scenario, we show the precise examples GradTail is upweighting in Figure \ref{fig:toy_example_class}. Here we separate the dataset into ``common," ``rare," and ``hard" by associating each of these categories with a specific range of normalized gradient dot product values. Because we chose the pivot value for upweighting in this setting to be 0, we associate rare examples with a small range around 0, or [-0.07, 0.07]. ``Hard" examples are any examples that fall below this range, and ``common" examples are above. As seen in Figure \ref{fig:toy_example_class}(a), most examples in the common class fall under ``common," while most examples in the uncommon class fall under ``hard." However, examples from \textit{both} classes are caught in the ``rare" range, and they form a fairly symmetric distribution around the true decision boundary. This result emphasizes that the definition of ``rare" or ``long-tail" that is most useful to model training can be at odds with the common semantic definition found in the literature; rather than just labeling infrequent classes as ``rare," we see ``rare" examples as high-impact examples that most help the model make correct predictions. These ``rare" examples can come from either the infrequent or the frequent class distribution, as long as they benefit predictive accuracy on more troublesome data.

\begin{figure}[ht]
\centering
    \subfigure[Hard Distribution Ground Truth]{
      \centering
      \includegraphics[width=0.3\linewidth]{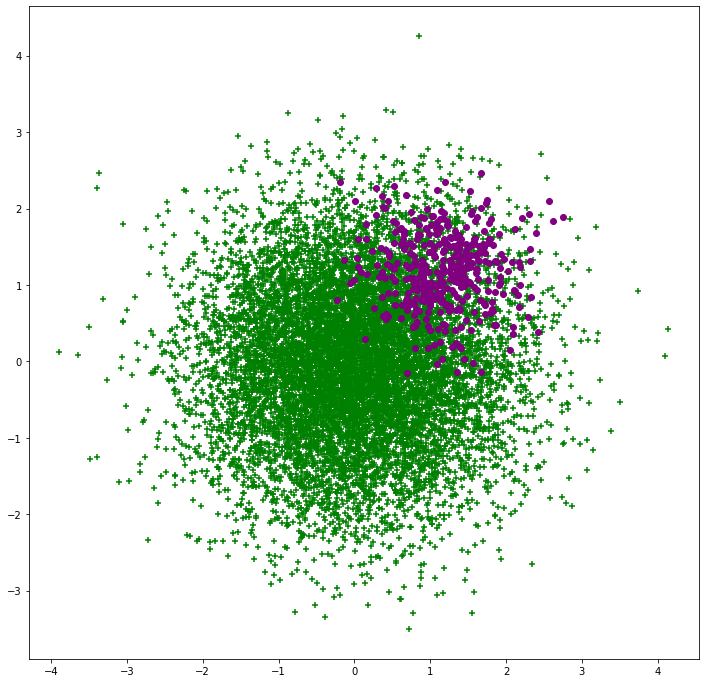}
    }
    \subfigure[GradTail]{
      \centering
      \includegraphics[width=0.3\linewidth]{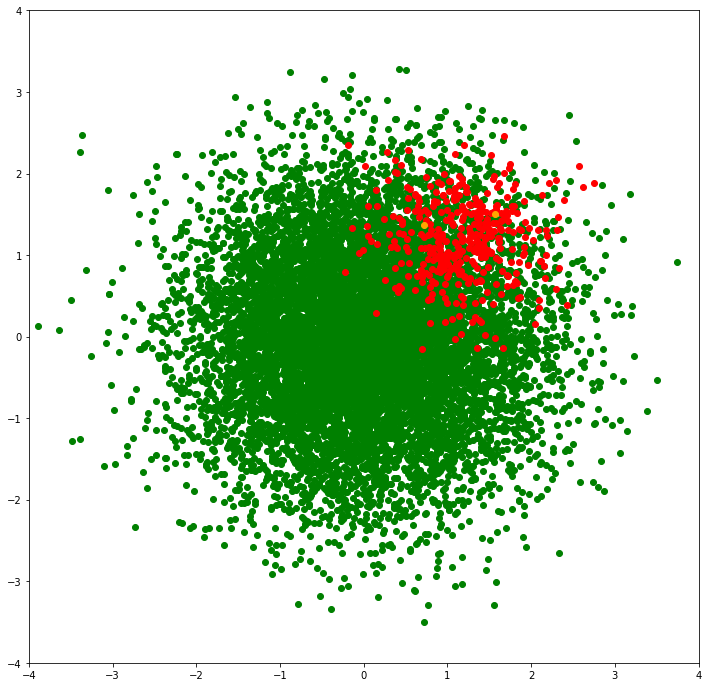}
    }
    \subfigure[Entropy]{
      \centering
      \includegraphics[width=0.3\linewidth]{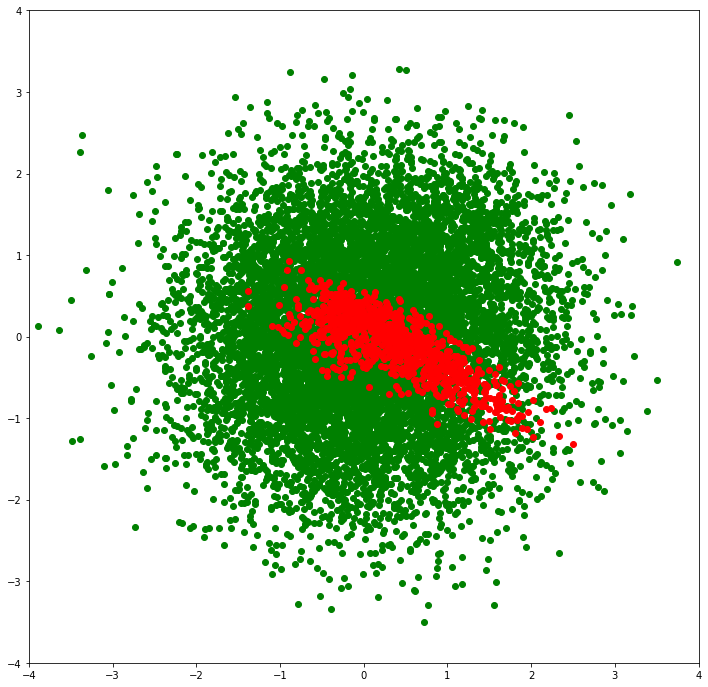}
    }
\caption{The same ``hard" vs ``rare" vs ``common" analysis as described in Figure \ref{fig:toy_example_class} but on a different distribution, as shown in (a). This harder distribution corresponds to the case where the common class has higher frequency than the uncommon class at every point in the input space. As shown in (c), entropy still produces incoherent results, and in (b) we observe that GradTail no longer outputs any examples within the ``rare" category. This result is sensible, as there are no high-leverage examples for which collecting similar examples will help us continuously improve the inferred decision boundary.}
\label{fig:toy_example_class_hard}
\end{figure}

As further illustration of these concepts, in Figure \ref{fig:toy_example_class_hard} we also investigate the interesting case when the uncommon class data is pulled far into the common class distribution, to the point where the common class frequency dominates the uncommon class frequency at all locations. In this setting, GradTail no longer outputs any data in the ``rare" category, but all of the uncommon class data is now labeled as ``hard." Because adding more uncommon data will not appreciably change the quality of the classifier in this case (unless we add a huge amount to put the class frequencies more into balance), the uncommon class data is purely ``hard," and will remain unaltered by GradTail.

We note that in both data distributions analyzed in Figures \ref{fig:toy_example_class} and \ref{fig:toy_example_class_hard}, plotting points of high entropy leads to an incoherent result. Entropy is a common metric within classification settings for example uncertainty, and is often used as a rarity classifier. Meanwhile, our GradTail methodology produces meaningful results even within the more difficult setting. 

It is pertinent to note that for our toy examples, although GradTail produces better class-normalized mean accuracy due to a significantly improved decision boundary, it lowers the total accuracy. GradTail moves the decision boundary but will end up misclassifying substantially more members of the common class as a result. This regression results from the fact that our toy example is very low-dimensional (deliberately so for ease of visualization), to the point where movement of the decision boundary is an explicit tradeoff between accuracy of one class over the other. We will show in the following sections that when we are in a very high-dimensional setting GradTail can improve overall accuracy as well. We hypothesize that the higher dimensionality reduces the possibility of having an explicit accuracy tradeoff between classes as there will always exist a perfectly separating hyperplane between different classes. 

\subsection{Classification}
\label{sec:imagenet}

\begin{table}
\caption{ImageNet Classification with GradTail. Higher numbers are better. All standard errors are within 0.05\%.}
\label{table:imagenet}
\vspace{-0.10in}
\begin{center}
\begin{small}
\begin{sc}
\begin{tabular}{lccr}
\toprule
Method & Accuracy (Top 1) & Accuracy (Top 5) \\
\midrule
Baseline    & 76.8& 92.9 \\
Focal Loss  & 76.2& 93.0\\
GradTail    & \textbf{77.2}& \textbf{93.1} \\
\bottomrule
\end{tabular}
\end{sc}
\end{small}
\end{center}
\vskip -0.1in
\end{table}

We describe within this section experiments on the Imagenet Large Scale Visual Recognition Challenge 2012 dataset, consisting of 1000 classes of various objects and other entities. For our experiments we use a ResNet-50 \cite{resnet} as our base network. Our GradTail model has a max loss weight of 3 and a pivot value of $0.0$. The GradTail gradient dot products are calculated on the last two layers, on both weights and biases.

\begin{figure*}[ht]
\centering
\includegraphics[width=1.0\linewidth]{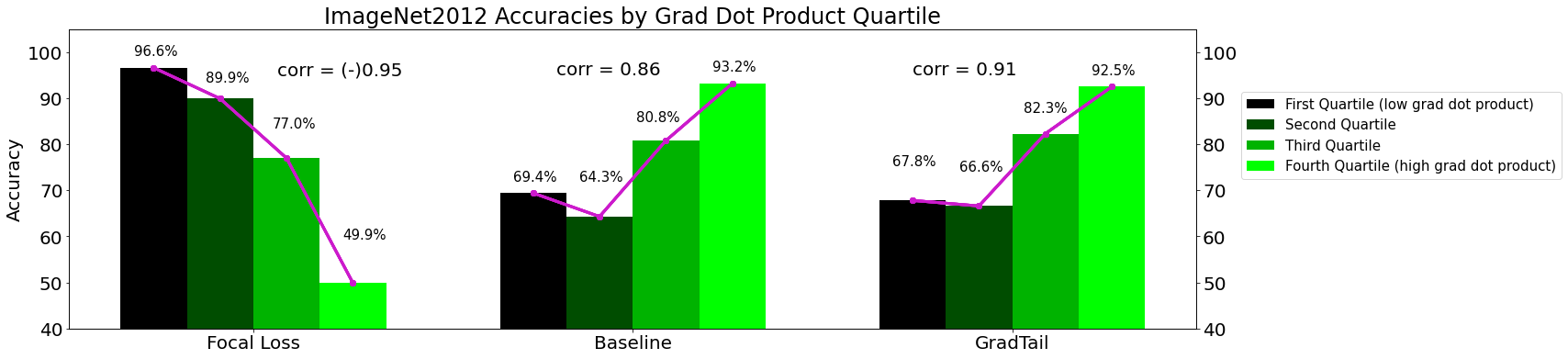}
\label{fig:quartiles}
\vspace{-0.25in}
\caption{ImageNet accuracies broken down by gradient dot product quartile. Lowest quartile corresponds to the lowest gradient dot products. Correlation coefficients between accuracy and gradient dot product value are also provided for each model.}
\end{figure*}

The main results are shown in Table \ref{table:imagenet} for a baseline model, the GradTail model, and a model with focal loss \cite{lin2017focal} used during training. We see that the top 5 accuracies are fairly clustered, but there is significant improvement within the GradTail model on top 1 accuracy. A closer look at the driving forces behind this improvement are encapsulated in Figure \ref{fig:quartiles}. Here we tabulate the top-1 accuracies for \textit{each quartile for the normalized gradient dot product}. The first quartile accuracies represent the accuracies amongst the subpopulation of data that lie in the lowest 25\% of the normalized gradient dot product value.

As expected, for the baseline we see that the accuracy tends to improve as the gradient dot product increases. However, this trend is reversed to a shocking degree when focal loss is used. Focal loss greedily upweights examples that produce higher loss, and the result is a complete decoupling between example accuracy and gradient dot product. Such a phenomenon is mathematically counterintuitive, as examples that disagree with the main gradient direction now completely drive the training. 

In contrast, GradTail tracks closely with the performance of the baseline on each of the quartiles, but beats the baseline handily in the middle two quartiles. Interestingly, the baseline performs better than GradTail in the first quartile, where hard examples of the highest aleatoric uncertainty lie. This effect may be driven by the baseline seeing higher loss and thus higher gradients for members of the first quartile during training. GradTail sacrifices some performance within the hardest quartile in favor of focusing on the more learnable middle quartiles. The overall effect is beneficial to the model. We also present in Figure \ref{fig:quartiles} the correlation value between gradient dot product band and performance, showing that GradTail produces the largest correlation. This increased correlation shows taht GradTail leads to \textit{better calibration} between gradient direction and performance, which we hypothesize is a desirable quality for a trained model.

The astute reader might observe that the GradTail dynamic weight profiles will be different for different model training runs, and therefore the quartiles in Figure \ref{fig:quartiles} are coupling to different examples for different baselines. While such analysis is correct, we deliberately used the quartile metric to emphasize our firm belief that \textit{being in the long-tail is not just a property of the data but also of the model as well}. Given that many rare examples are borderline learnable by baseline models and therefore might happen to be learned well on certain runs, having model dependence as part of our long-tail definition affords us additional flexibility in identifying the true problem points that we can try to tune our models to learn.

\begin{figure*}[ht]
\centering
    \subfigure[Easy Examples (high gradient dot product)]{
      \centering
      \includegraphics[width=0.8\linewidth]{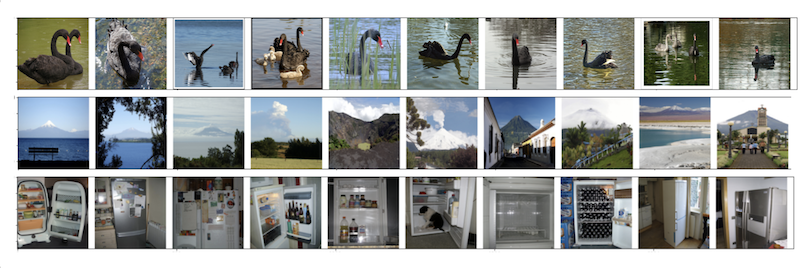}
    }
    \linebreak
    \subfigure[Hard Examples (low gradient dot product)]{
      \centering
      \includegraphics[width=0.8\linewidth]{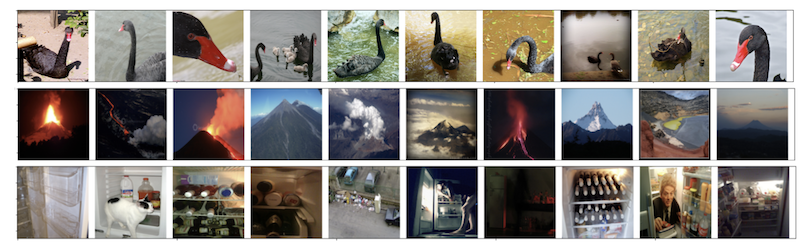}
    }
\caption{Gradient dot product visualization on the ImageNet dataset. Gradient dot products are averaged over 12 saved models around ~10\% through the total training regimen. The examples within the test set that exhibit both the highest (a) and lowest (b) gradient dot product for the classes of black swan (top row), volcano (middle row), and refrigerator (bottom row) are visualized. High gradient dot product implies that an example is generally aligned with the average training direction and thus is in-distribution, while low gradient dot product points to a difficult example that is out-of-distribution. In all three class cases, the hard examples are semantically reasonable and exhibit image features that make them more difficult (e.g. murky water for swans, eruptions for volcanos, and odd lighting or other animal/humans present for refrigerators).}
\label{fig:imagenet_vis}
\end{figure*}

Despite our confidence in the metrics, it still would be reassuring to know that the gradient dot product we calculate as part of GradTail produces semantically meaningful results (as we saw for our toy example in Section \ref{sec:toy-example}). In Figure \ref{fig:imagenet_vis} we visualize examples in three randomly chosen ImageNet classes that produce the highest and lowest gradient dot product values. We observe that for all three classes, high dot product examples are semantically similar while low dot product examples exhibit odd features. For volcanos, mid-eruption volcanos usually produce low gradient dot product, which is sensible as most volcanos within the dataset are dormant. Swans that exhibit low gradient dot product are often on land or in murky water, which hints that the environment is salient to the model for swan classification. Refrigerators with low gradient dot product often have additional agents in the scene (e.g. pets or humans), or are shot at odd angles or distances. In general, all of these visualizations improve our confidence that GradTail hones in on the appropriate examples. 

It is crucial to note that the visualizations in Figure \ref{fig:imagenet_vis} were all generated by models at around 10\% through the training regimen. Unlike conventional long-tail work in the literature, our GradTail algorithm is fairly discriminative early on in the training, which demonstrates that model convergence does have to be a requirement in identifying and mitigating rare examples within the dataset. 

\subsection{Dense Regression (Depth Estimation)}
\label{sec:wod}

For our regression problem, we choose camera-based monocular depth estimation on the the Waymo Open Dataset. The dataset consists of camera images downsampled to $192\times 480$ paired with ground truth that is per-pixel depth measurements from LiDAR. Our model uses a 5-block MobileNet \cite{mobilenet} encoder with a 5-layer decoder that also concatenates encoder block outputs in a similar way as U-Net \cite{unet}. The filter layouts in the encoder network were searched via NAS \cite{tunas} in the baseline setting to maximize potential performance of the model, and kept fixed for all experiments.

Long-tail mitigation on dense regression data is a particularly exciting topic, as long-tail work has largely been restricted to single-output classification problems. However, there is every reason to believe that regression problems and dense prediction problems are all hurt by issues in the long-tail. As gradients are universal quantities within deep learning training dynamics, our proposed methodology is well-suited to tackle dense regression with minimal modifications from the classification setting.

The density of the output for our chosen problem can be one potential source of inefficiency. As we have to calculate a gradient dot product for each output, and our outputs are now in a grid of over \SI{92000} pixels, it is not recommended to perform this calculation for each pixel. Even if it were compute-efficient to do so, there is significant spatial correlation between adjacent pixels and so it would be a waste of compute to consider long-tailed upweighting at every pixel location. Instead, we take six random patches of randomly sampled sizes between $20\times 20$ and $100\times 100$ and randomly sampled locations within the full image space. We also take a seventh patch consisting of any pixel that was not included in any proposal to ensure that we backpropagate some error signal at every location. We then mean the loss in each patch and concatenate for an effective batch size of $7\times$ the original batch size.

\begin{table*}[t]
\caption{Waymo Open Dataset Monocular Depth Estimation with GradTail. The lower the better for all metrics within this table. Total MRE has standard error within 0.03\%, while all other results have standard error within 0.05\%.}
\label{table:depth}
\vspace{-0.12in}
\begin{center}
\begin{small}
\begin{sc}
\begin{tabular}{lcccccr}
\toprule
Method & MRE $<$20m (\%) & MRE 20-40m (\%) & MRE 40-60m (\%) & MRE $>$60m (\%) & Total MRE (\%) \\
\midrule
Baseline    &\textbf{10.1}&11.9&14.2& 16.8& 11.0 \\
GradTail, pivot $-2.5$   &10.5&11.9&14.2& 16.8& 11.2 \\
GradTail, pivot $+0.5$   &10.5&11.9&13.9& 17.2& 11.3 \\
GradTail, pivot $-0.5$   &\textbf{10.1}&\textbf{11.7}&\textbf{13.7}& \textbf{16.5}& \textbf{10.8} \\
\bottomrule
\end{tabular}
\end{sc}
\end{small}
\end{center}
\vskip -0.1in
\end{table*}

The results of our experiments on depth estimation are shown in Table \ref{table:depth}. We see that the overall error rate improves by a marginal (though statistically significant) amount, but the picture is clearer when we tabulate the error rate for different depth ranges. The error rate improves significantly for points within the 40-60m range, while being minimal for close points in the 0-20m range. Physically, we know that points further in the distance are necessarily less common within the dataset (due to parallax). Unlike the ImageNet case (Section \ref{sec:imagenet}), where example rarity is semantically informed and thus complicated, rarity dependence on distance within depth estimation datasets is a physical property of the sensors and thus a reasonable proxy for long-tailedness. These results are therefore a reassuring signal that although the overall improvement is small, we substantially improve performance in areas where we lack data and which generally are problematic for standard models.

The pivot parameter is also clearly important in this scenario; a pivot close to zero but slightly negative provided the best results. Positive pivots, which would upweight in-distribution examples, and high negative pivots, which would upweight only hard examples, both degraded model quality. The optimal pivot being close to zero further supports our emphasis on orthogonal gradients as meaningful. 

\section{Discussion}

Our goal for introducing GradTail is twofold: first, we want to demonstrate it as a general method that works to mitigate performance issues on long-tail parts of the data distribution. We showed through substantial analysis on a low-dimensional toy example that GradTail works well within the context of highly lopsided data, and produces a sensible decision boundary even when one data distribution is $25\times$ the frequency of the other. We also showed improvements in performance on rare segments of the data in both a classification (ImageNet) and dense regression (depth estimation on Waymo Open Dataset) context. The latter is especially exciting, as long-tail methods have largely been reserved for classification settings due to the complexity of moving to a continuous output space. However, because gradients are well-suited to continuous problems and are universal to neural network training, they are powerfully general entities to use for regression long-tail mitigation.

Our second goal is to challenge pre-existing notions of what long-tailedness and rarity within a dataset mean. Long-tailedness has largely been a work of manual labor in the literature, with humans semantically labeling objects to be rare and ad-hoc triaging when new rare classes crop up. Datasets like iNaturalist \cite{dataset1} exist that boast long lists of long-tailed classes, where long-tailedness is a property of a class rather than an example. While such analysis has been crucial in robustifying neural network thus far, we hope that our work provides a fresh look at how we can improve model training by reasoning about long-tailedness at a learned, granular level and using model dynamics as our fundamental signal. 

\subsection{A Note on Compute/Memory Costs}
Throughout this work, we noted that Gradtail requires a gradient dot product to be calculated for each example. Although this may seem expensive, example-level gradient computation is already done in the vast majority of models, which allows us to reuse calculations to implement GradTail. The overhead then becomes minimal, with the most expensive step becoming the taking of a single dot product.

However, we have found that setting up this reused computation is difficult in most modern deep learning frameworks, as it is common within such frameworks to hide the backwards pass deep within the system backend. Intercepting the computation right before the per-example gradient signal is summed across the batch dimension can then become challenging. Up until now, such design decisions by these frameworks may have been due to the limited use of intercepting such a computation. However, we hope that our work on GradTail and followup research will clarify the potential benefit that access to the per-example gradients can offer, and that we can encourage designs of deep learning framework that allow better interfacing to the gradient computation backend. 

\section{Conclusions}

We presented GradTail, a gradient-based dynamic weighting algorithm that upweights examples during training based on their rarity. We showed how GradTail works to produce a reasonable decision boundary on an extremely lopsided low-dimensional classification problem, as well as working to mitigate poor performance on rare examples within a higher-dimensional classification problem (ImageNet). Crucially, we show that GradTail generalizes to dense regression settings as well, which have hitherto been relatively inaccessible to long-tail methods. Ultimately, we see GradTail as an important tool within the toolkit of any deep learning practitioner, but also see it as significant evidence that there is much to be learned about any complex data distribution from the information-rich but oft-ignored dynamics of training a model. It may be that to truly robustify our models, we need to focus not on where our models converge, but rather on the myriad twisting paths by which they get there.

\clearpage

\bibliography{example_paper}
\bibliographystyle{icml2022}

\newpage
\appendix
\onecolumn
\setcounter{figure}{0}
\renewcommand\thefigure{A.\arabic{figure}}
\section{Discussion on Aleatoric vs Epistemic Uncertainty and Gradient Dot Products}
\label{sec:uncertainty-discussion}

Within the main body of this work, we repeatedly make the claim that orthogonal (or close-to-orthogonal) gradients with respect to the average gradient indicate examples with high epistemic uncertainty, while gradients with high negative gradient dot products indicate examples with high aleatoric uncertainty. We would like to take the opportunity to further discuss the connection between our work and these classical concepts in uncertainty estimation.

Traditionally, epistemic uncertainty is also known as \textit{model uncertainty}, and reflects our model's lack of knowledge to learn a certain piece of data. Although there are various ways of modeling such uncertainty (e.g. putting a prior on the model weights as done in Bayesian deep learning \cite{kendall2017uncertainties}), a key property of epistemic uncertainty is that it can always be alleviated by collecting more data. That key property immediately creates a fundamental link between epistemic uncertainty and long-tailedness. In contrast, aleatoric uncertainty details irreducible noise within the data that will be present regardless of how much data we collect.

We thus ask ourselves what kinds of data a sufficient model will learn better given more examples of that data type. We argue that this is the point where re-evaluating the problem in the context of model gradients becomes especially helpful. Namely, from basic calculus there is at least a local guarantee that for a forward-pass model $M$, should an example $x$ with label $y$ produce gradients $\nabla_wL(M(x); y)$ that produce dot products that are zero or greater with respect to the average gradient, then applying these gradient updates will: 

\begin{enumerate} 
\item Reduce the loss $L(M(x); y)$.
\item Not degrade the performance of the average example within the dataset.
\end{enumerate}

Thus, we note that \textit{collecting more data with a similar gradient as $x$ and therefore a similar dot product will necessarily reduce the uncertainty of $x$ within the model $M$ trained on the full dataset}, while not reducing the performance of the model otherwise. This argument tells us that the uncertainty of data point $x$ is largely epistemic.

In contrast, if $\nabla_wL(M(x); y)$ is antiparallel or produces negative dot product when dotted with the average gradient, then collecting more examples with a similar gradient profile will hurt the performance of the average example for that model. Thus, as we train on the \textit{full dataset}, although we may overfit to the example $x$ and reduce its uncertainty, the overall model performance will degrade and become even more susceptible to the noise of misbehaving examples such as $x$. Such behavior does not fulfill the key property of epistemic uncertainty of being always reducible with more data collection, and so we associate its poor performance more with high aleatoric uncertainty. 

We note that much of this argument involves contextualizing sample uncertainty within a larger ecosystem of model training. In our view, the uncertainty of any given example is only meaningful in the context of a model that is making inference on all that data, as we never observe training examples by themselves in a vacuum. Such considerations are a key driving force of our proposed GradTail algorithm, as the dynamic quality of the algorithm ensures that the computed uncertainty (via gradient dot product) of a given example will often change with time and depends on the state of the model. We find that reasoning about uncertainty while tied to a model state is crucial for determining not only examples of differing uncertainty types, but examples that will be of optimal practical use to model training. Empirically, this claim is well supported, especially by our visualizations in Figure \ref{fig:toy_example_class} where we showed that long-tail examples cluster around the decision boundary, and in Table \ref{table:depth} where we showed that performance degrades for our algorithm for large negative pivots.

\section{Training Details}

Unless otherwise noted, all trained models were trained on TPU cores within the TensorFlow framework. The Toy Example and depth estimation experiments were performed on TensorFlow 1, while the ImageNet experiments were performed on TensorFlow 2.

\subsection{Toy Example}
\label{sec:toyex-appendix}

The toy example network is a simple MLP model with one hidden layer of 5 neurons. The network is barebones and does not employ any of the standard tools like batch normalization or dropout. Training for all models is performed for \SI{10000} steps with an initial learning rate of \SI{1e-4}. The learning rate is not decayed. We optimize using a look-ahead momentum optimizer \cite{sutskever2013importance} with momentum \SI{0.9} .

Our GradTail layer uses all weights (both convolutional weights and biases) within the MLP for comparison, and produces a maximum upsampling weight of \SI{15} .

\begin{figure}[ht]
\centering
\includegraphics[width=1.0\linewidth]{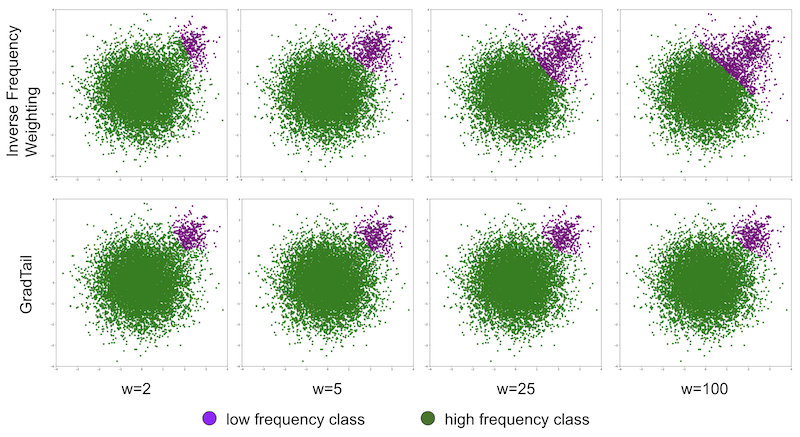}
\label{fig:frequency_weighting}
\caption{GradTail classification versus Inverse Frequency Weighting classification on an imbalanced toy dataset. The top row corresponds to inverse fequency weighting results while the bottom are GradTail results. Each column corresponds to a different weighting. For inverse frequency weighting this is the weight assigned to training examples of the uncommon class. For GradTail this is the maximum upweight factor for an example that is within the long-tail.}
\end{figure}

We also compare GradTail with inverse class-frequency weighting in Figure \ref{fig:frequency_weighting}. By inverse class-frequency weighting, we mean that during training, examples within the common class are assigned weight \SI{1} and examples within the uncommon class are assigned weight $w\geq 1$. This weighting is a common technique used in practice to deal with known class imbalance within the training dataset. We compare this to the GradTail model, for which $w$ is the maximum weight assigned to a detected long-tail example. The baseline $w=1$ case is the one shown in Figure \ref{fig:toy-example}b.

We see that as $w$ increase, inverse frequency weighting forces the decision boundary further down and to the left, reflecting the increased leverage that the uncommon-class examples now have on the training dynamics. If we set the frequency weighting to the exact frequency ratio between the common and uncommon classes (in this case that ratio would be 25), the resulting decision boundary is rather aggressive and encroaches very far into the common class territory. Finding the right balance becomes a hyperparameter search problem, which is exponentially exacerbated when there are more than two classes.

In contrast, GradTail is consistent throughout regardless of what we set as the maximum upweighting factor. We attribute this robustness to the dynamicness of the method; GradTail cannot deviate from a good balance point between the classes because once one class starts to dominate, the other class's examples are more likely to be detected as long-tail. GradTail thus provides a \textit{restoring force} to the system, leading to a stable convergence that would be elusive to manual methods like inverse frequency weighting where the same weight is applied throughout the training run regardless of how the model trains.

We further note that although inverse frequency weighting is of some utility in some scenarios, one of our core beliefs within this work is that we need to move away from a manual semantic definition of long-tail. An example is not necessarily in the long-tail just because it belongs to a less common class, as the class definitions themselves may come from semi-arbitrary semantic labels and highly overlapping generating distributions. Overfitting our methodologies to a semantic class-based definition of long-tail may lead to subpar performance, as demonstrated here.

\begin{figure}[ht]
\centering\centering
    \subfigure[Refrigerator]{
      \centering
      \includegraphics[width=1.0\linewidth]{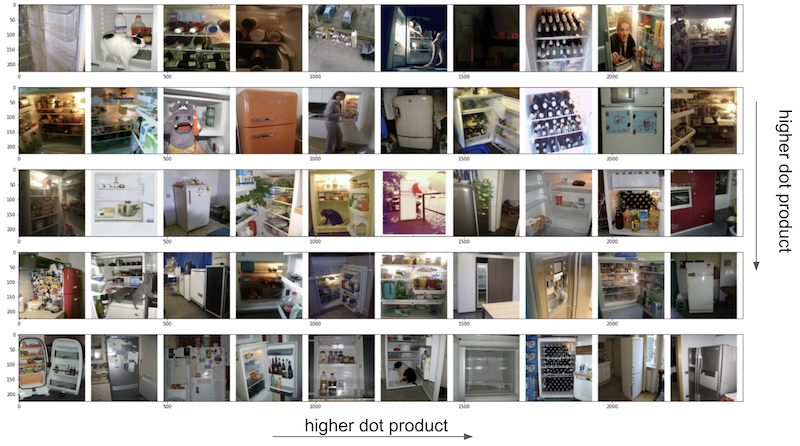}
    }
    \linebreak
    \subfigure[Volcano]{
      \centering
      \includegraphics[width=1.0\linewidth]{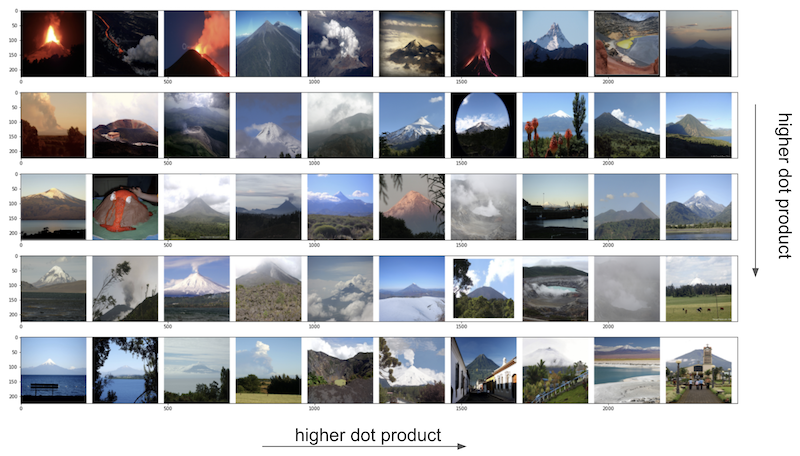}
    }
    
\caption{ImageNet visualization ordered by gradient dot product for two classes within the ImageNet2012 test set. Images produce higher gradient dot product with the mean gradient as the images go from left to right, and then top to bottom. Images on the bottom right of each category have the highest gradient dot product, while images on the top left of each category have the lowest.}
\label{fig:imagenet_vis_full}
\end{figure}

\subsection{ImageNet}

ImageNet inputs are downsampled to 160x160 inputs before being put through a standard ResNet v1 architecture \cite{resnet}. Models are trained for 1.7 million steps with an initial learning rate of 0.025 and a cosine learning rate decay profile \cite{loshchilov2016sgdr}. We use the same optimizer as in the toy example (look ahead momentum with momentum parameter 0.9), and also use a label smoothing of 0.1. We set regularization at 6e-5 and set batch size to 256.

Our GradTail layer uses gradients from the final two layers (both convolutional weights and biases), and the maximum GradTail weight is set to 3. We use a pivot value of 0 for all our ImageNet experiments, which is consistent with our interpretation that orthogonal gradients belong to long-tail examples. 

\subsection{Waymo Open Dataset}

We perform monocular depth estimation on the camera images of the Waymo Open Dataset by training a regression model to produce metric per-pixel depth predictions from camera image only. The depth ground truth was obtained by projecting synchronized LiDAR points into the respective images. The model input and output resolution is set to $192\times 480$, and images are fed through a 5-block MobileNet~\cite{mobilenet} encoder with a corresponding decoder and skip connections. The training loss is a simple $\ell_1$ loss applied only to pixels that have a ground truth depth value assigned to them. The learning rate is kept constant at $\textit{lr}=0.0002$ throughout the training, and the batch size is set to $10$.

Our GradTail layer uses just the biases in the first two upsampling layers within the decoder, which allows our method to be especially efficient within this setting. The maximum GradTail weight is set to 15. 

\section{More Insight on Methodology Hyperparameters}

Our proposed GradTail algorithm is relatively simple and straightforward, but does come with a few hyperparameters. In our experience, GradTail is fairly robust (see for example the frequency weighting experiments in Section \ref{sec:toyex-appendix}, but we include many of these hyperparameters as guards against edge cases and specific scenarios where dynamic drift might be severe enough to throw off GradTail without further mitigation. In this section we go through the hyperparameters and offere a few additional insights on each of them. 

\textbf{Pivot parameter $p$}. The pivot parameter $p$ is the main hyperparameter and the one that has the largest influence on training (see for example the dense depth estimation results in Section \ref{sec:wod}). The pivot parameter represents our belief for what examples count as "long-tail" and thus must be upweighted. Without any additional information, we recommend always setting the pivot to zero for initial experiments. This setting is because a pivot of zero means we upweight examples that backpropagate orthogonal gradients, which represent examples that are learnable but still difficult due to lack of model exploration of the parameter space. However, we leave the pivot as a hyperparameter because small swings in the negative and positive direction can offer additional (or less) regularization, with more negative pivots resulting in more regularization. Because we ultimately want to optimize our models on a test set, such regularization potential can be useful. 

\textbf{Activation function $f$}. The activation function transforms a normalized gradient dot product into a sample weight, which should be at least 1.0. We chose a sigmoid-like activation function as it provides us with a sharp peak close to the desired pivot, which allows us to be more selective with what samples to upweight. 

\textbf{Decay rate $\lambda$}. This is the decay rate by which we assign new average gradient values to the exponential moving average and variance $\sigma$. We set this decay rate in our experiments to 0.99, which means that our pool of average gradients reflects on average the gradients of the last 100 batches. 

\textbf{Variance $\sigma$}. Although not a hyperparameter, the zero-centered variance $\sigma$ is worth discussing as it is an additional normalization term that may seem mysterious at first. We add $\sigma$ into the main algorithm because our algorithm is a dynamic one which deals with moving statistics within a network that is training. In many cases the drift of $\sigma$ through training was only mild, so it is plausible to use GradTail without $\sigma$, but as a safety measure it is still recommended.

\section{Full ImageNet2012 Visualization for 2 Classes}

In the main paper, we showed the highest and lowest dot product images for some ImageNet classes in Figure \ref{fig:imagenet_vis}. For completeness, we include all the images within the test set (50 each) for two ImageNet classes, volcano and refrigerator, in Figure \ref{fig:imagenet_vis_full}. 

For each class, the gradient dot products for each image become higher from left to right in each row, and then from top to bottom. In other words, the top left image in each class produced the lowest gradient dot product, while the bottom right image in each class produced the highest. In each case, zero gradients occur close to the beginning of the third row.

We see that as gradient dot products become higher and higher, there is a clear semantic trend for images to standardize into fairly similar images. For refrigerators we begin to see images of full refrigerators (with doors open or closed) without much debris or extraneous elements in the frame. For volcanos we see many images of clear skies and snow-capped peaks. At the lower end of the gradient dot product scale, we begin to see odd features like animals or humans in front of a refrigerator or volcanos that are mid-eruption. 

We emphasize, however, that although we include these visualizations for completeness, we do not fully recommend the practice of trying to semantically reason why any given data input may or may not be difficult for a given model. One of our major intended contributions of our work was to demonstrate that it is reasonable to define long-tailedness purely as a function of the data and model, without reference to human defined classes or class hierarchies (which are always susceptible to at least a bit of arbitrariness). Although it is a useful sanity check to see that there is a clear semantic difference in ImageNet data as we progress along the dot product scale, we hope to emphasize that it is more useful to think about long-tailedness in terms of parameter space exploration and learnability, both of which our gradient framework addresses explicitly. 


\end{document}